\documentclass[10pt,journal,compsoc]{IEEEtran}

\usepackage[utf8]{inputenc}
\usepackage[T1]{fontenc}
\usepackage{graphicx}
\usepackage{booktabs}
\usepackage{array}
\usepackage{xcolor}
\usepackage{hyperref}
\usepackage{balance}
\usepackage{tabularx}
\usepackage{multirow}
\usepackage{caption}
\usepackage{tabularx}
\usepackage{booktabs}
\usepackage{array}
\usepackage{amsmath}
\usepackage{amssymb}

\usepackage[table]{xcolor}
\usepackage{tabularx}
\usepackage{booktabs}
\usepackage{array}
\usepackage{float}
\definecolor{safetyred}{RGB}{255,235,235}

\usepackage{tcolorbox}
\usepackage{tabularx}
\usepackage{booktabs}
\usepackage{array}
\newcolumntype{Y}{>{\raggedright\arraybackslash}X}

\definecolor{safetyred}{RGB}{255,240,240}
\definecolor{headerblue}{RGB}{43,87,151}

\hypersetup{
    colorlinks=true,
    linkcolor=blue,
    citecolor=blue,
    urlcolor=blue
}

\begin{document}

\title{Personal Care Utility: Health as Everyday Infrastructure}

\author{Mahyar~Abbasian, Elahe~Khatibi, Saba~A.~Farahani, Nitish~Nagesh, Arshia~Ilaty, Hooman~Sajjadi, Amir~Rahmani, and~Ramesh~Jain
\IEEEcompsocitemizethanks{
\IEEEcompsocthanksitem All authors are with the University of California, Irvine.\protect\\
E-mail: \{abbasiam, ekhatibi, fazizaba, nnagesh1, ailaty, hsajjadi, a.rahmani, rcjain\}@uci.edu
\IEEEcompsocthanksitem Corresponding author: M. Abbasian (e-mail: abbasiam@uci.edu).
}}

\IEEEtitleabstractindextext{
\begin{abstract}
Healthcare is essential, expert, and episodic by design — built around the roughly one hour per year a person spends with a clinician. The 8,759 hours outside clinical settings, where eating, sleeping, movement, medication, and stress actually shape long-term health, have no comparable infrastructure. The bottleneck for personalized health is not raw data or reasoning capability; it is the absence of that infrastructure layer.
This paper introduces the Personal Care Utility (PCU): a layered, event-driven architecture proposed as the missing utility for everyday health, in the way that payments, networks, and power are utilities for their domains. PCU organizes continuous personal signals into semantically meaningful life events through a Personicle, estimates dynamic health state against personal baselines, reasons about cause and context, and routes guidance through an orchestrator that separates clinical decision logic, behavioral strategy selection, and natural-language expression. This separation lets large language models support reasoning and communication while keeping safety-critical clinical decisions grounded in validated evidence.
We instantiate PCU for Type 2 Diabetes — turning CGM, meal, activity, medication, sleep, stress, and clinical data into glycemic events, individualized state estimates, causal explanations, and knowledge-grounded interventions. A day-in-the-life scenario shows the same infrastructure producing real-time nudges, weekly summaries, medication check-ins, silence, or deterministic safety alerts depending on context and risk. We close with how PCU generalizes to other chronic conditions and the governance questions any always-on personal health utility must address. The result is a blueprint that treats personalization not as a final messaging layer, but as an architectural property of everyday health guidance.
\end{abstract}
}

\maketitle

\IEEEdisplaynontitleabstractindextext

% ================================================================
\section{The 8{,}759-Hour Gap and a New Opportunity}
% ================================================================
An Intensive Care Unit is a marvel of continuous computing. Sensors track every vital sign, algorithms flag anomalies in real time, and expert teams orchestrate responses with precision. Yet this vigilance evaporates the moment a patient walks out the hospital door. By any reasonable accounting, an individual spends the vast majority of their year---of order $8{,}759$ hours---outside clinical settings, eating, sleeping, working, and worrying, making the cumulative decisions that shape their long-term health far more than any single consultation~\cite{topol2019highperf,flores2013p4}. Our computing systems, however, remain largely fixated on the few clinical visits — an average of about one hour per person per year — leaving most of life unmonitored, uninterpreted, and unsupported.

This is not merely an inconvenience. Chronic conditions---diabetes, hypertension, heart failure, depression---are managed almost entirely within those unsupported hours, shaped by behavior, environment, circadian rhythms, and social context~\cite{topol2019highperf,nag2019health}. Clinical care is essential, expert, and episodic by design — a structural choice, not a flaw, but one that leaves the 8,759 hours between encounters without an infrastructure of their own. Despite a steady expansion of wearable sensors and mobile health applications, today's digital health ecosystem remains fragmented: data are siloed across incompatible platforms, individuals receive raw metrics rather than contextual synthesis, and few systems adapt based on outcomes~\cite{flores2013p4,hood2011p4,nag2019health}. The result is a familiar paradox: more data than ever before, but often less actionable insight than a five-minute conversation with a knowledgeable clinician.

Several converging developments have changed what is now possible. Continuous physiological sensing has matured, foundation models have begun to learn rich representations of individual trajectories, large language models have become capable reasoning and communication interfaces, and rigorous frameworks for timed behavioral support have emerged from the pervasive-computing community---each, however, with characteristic limitations when deployed alone. The point here is simpler: the bottleneck of personalized health is no longer raw data acquisition or raw reasoning capability; it has shifted to data interpretation, contextual reasoning, and the delivery of guidance that is sensitive to who the person actually is~\cite{powell2024biomarkers,abbasian2025pcu}.

In this paper we argue that what is missing is not more data, not more AI, and not more apps in isolation. What is missing is the connective tissue: a system that perceives the person through heterogeneous personal data, remembers their history and preferences, reasons about cause and context against their own baseline, draws on validated knowledge, and acts by delivering guidance with the right tone, at the right moment, through the right channel. We describe this connective tissue as the \textit{Personal Care Utility} (PCU)---an always-on AI agent for everyday health, layered and event-driven by design~\cite{abbasian2025pcu,nag2019health,rahmani2022mental}. The choice of the word utility is deliberate: every modern domain — electricity, water, payments, networks — has matured by acquiring a continuous, person-centered infrastructure layer beneath the institutional services that depend on it. Health has never had its own such utility; PCU is proposed as that layer. Built as public infrastructure where such infrastructure exists, PCU is designed to be the person-centered layer that institutional health services participate around, rather than another silo of its own.  Like a GPS that navigates differently for every driver based on destination and real-time conditions, the PCU is meant to navigate health differently for every individual, drawing on continuous personal signals while remaining firmly bounded by clinical evidence and safety guarantees.
% ================================================================
% ================================================================
\section{A Convergence Worth Structuring}
% ================================================================

The PCU does not appear in a vacuum. It builds on and integrates several lines of work that have each made important progress on parts of the problem of everyday personalized health. We briefly review five such strands---spanning sensing, personalization, personal representation, intervention timing, and conversational reasoning---and then describe what the PCU takes from each.

\textbf{Mobile health and continuous personal sensing.} The data substrate outside the clinic has expanded dramatically. Over-the-counter continuous glucose monitors~\cite{ada2025,battelino2019tir}, clinically scrutinized wrist-worn wearables for sleep, heart rate, and activity~\cite{dhingra2023wearables,nagappan2024wearables}, smartphone-derived passive digital biomarkers~\cite{insel2017digital,powell2024biomarkers}, and foundation models trained directly on physiological streams~\cite{gluformer2025} together provide dense longitudinal signal on how individuals actually live. The acquisition problem is largely solved; the interpretation problem is not. Signals remain siloed, raw metrics are surfaced without context, and few systems convert this substrate into individualized, intervention-aware guidance. The PCU treats the substrate as a given and locates its contribution above it.

\textbf{Personalization as empirical necessity, not stylistic choice.} A second body of evidence shows that the substrate above only becomes useful when interpreted against the individual rather than the population. Long-standing work on heterogeneity of treatment effects documents that the average response measured in randomized trials often masks substantial benefit for some, little for many, and harm for a few~\cite{kravitz2004hte}, a pattern now confirmed across chronic disease domains as wearable and continuous-sensing studies reveal large interindividual variation in everyday physiology and in response to lifestyle and pharmacologic interventions~\cite{tian2023wearablemetabolic,zeevi2015personalized}. The methodological response---N-of-1 and single-patient trial designs that treat each individual as their own reference~\cite{kravitz2021nof1}---operationalizes at the trial level what the PCU adopts at the system level: the average response is a poor description of any specific person, and any always-on system that ignores this will systematically underperform. The PCU therefore interprets every detected event against personal baselines rather than population norms.

\textbf{Personal models, digital twins, and personal knowledge graphs.} Representing the individual has been pursued from several directions. Digital twins for health have evolved from physiological replicas toward multi-scale, behaviorally-aware representations~\cite{vallee2026msdt,katsoulakis2024dt4h,naik2025humandt}, though scoping reviews note that few deployed twins are simultaneously personalized, dynamic, and predictive~\cite{naik2025humandt}. In parallel, patient-centric knowledge graphs encode an individual's clinical history, observations, and preferences in structured queryable form~\cite{skjaeveland2024pkg,alkhatib2024pckg,jiang2024graphcare,sarani2024dtkg}, and causal memory architectures such as REMI attach causal schemas on top, grounding recommendations in inferred individual cause--effect structure rather than mere similarity~\cite{raman2025remi}. A complementary thread treats the person as a chronological stream of life events: the \textit{Personicle}~\cite{jalali2014personicle,jalalijain2021eventmining} and the \textit{Objective Self}~\cite{jain2014objectiveself} represent daily experience as a computable biography of episodes rather than a static record. The PCU adopts the shared intuition behind these efforts.

\textbf{Just-in-time adaptive interventions and the timing of support.} A strand with deep roots in this venue concerns when continuous data should be turned into action~\cite{spruijt2014dynamic,nahumshani2018jitai}. The Just-in-Time Adaptive Intervention (JITAI) framework specifies that support is most effective at moments of opportunity or vulnerability when the person is receptive~\cite{nahumshani2018jitai,nahumshani2026jitai}; empirical work shows that adapting timing alone substantially improves engagement~\cite{mishra2021receptivity,pielot2017opportune}, and microrandomized trials offer rigorous methodology for evaluating per-decision personalization~\cite{klasnja2015microrandomized}. A complementary thread reminds us that excessive notification is counterproductive and that \textit{not} delivering a message can itself be a designed action. The PCU adopts this directly, treating silence as a first-class orchestration outcome.

\textbf{Conversational health agents and personal-health LLMs.} Most recently, large language models have emerged as reasoning and communication interfaces for health. Frontier models can encode clinical knowledge and reach expert-level performance on standardized medical question answering~\cite{singhal2023llmsmed,singhal2025expertqa}, conduct multi-turn diagnostic dialogue matching primary-care physicians under controlled conditions~\cite{tu2025amie}, and deliver expert-grade coaching from longitudinal wearable data~\cite{cosentino2025phllm}. Tool-augmented agents have begun to extend these capabilities to longitudinal personal data, with systems such as openCHA~\cite{abbasian2025opencha,rahmani2022mental} and the Personal Health Insights Agent~\cite{merrill2024phia} showing that LLMs can reason meaningfully about wearable streams when given the right scaffolding. A consistent pattern emerges: monolithic LLMs are increasingly capable but insufficient on their own, and performance, safety, and personalization all improve when the model is embedded in a structured pipeline with appropriate data, tools, and constraints~\cite{mehandru2024agentsclinic,omar2025llm}. The PCU is one such structured pipeline. It uses LLMs for reasoning and communication, while an agentic architecture grounds clinical decisions in validated evidence and keeps them auditable.

\textbf{Where PCU sits.} Each strand above contributes something the others lack: sensing substrate, individualized interpretation, persistent personal representation, principled timing, or fluent reasoning. None, on its own, delivers an integrated system. Concurrent proposals such as the ``anatomy of a personal health agent''~\cite{heydari2025pha} reflect the same recognition. The PCU is one such integrated proposal---an event-driven, agentic system that brings these strands together while preserving the safety and equity considerations any always-on health system must respect~\cite{veinot2018goodintent,pfohl2024equitymedqa,sabben2024access,rodriguez2025langdisp}.

% ================================================================
\section{PCU Architecture: A Brief Overview}
% ================================================================

\begin{figure*}[t]
    \centering
    \includegraphics[width=0.95\textwidth]{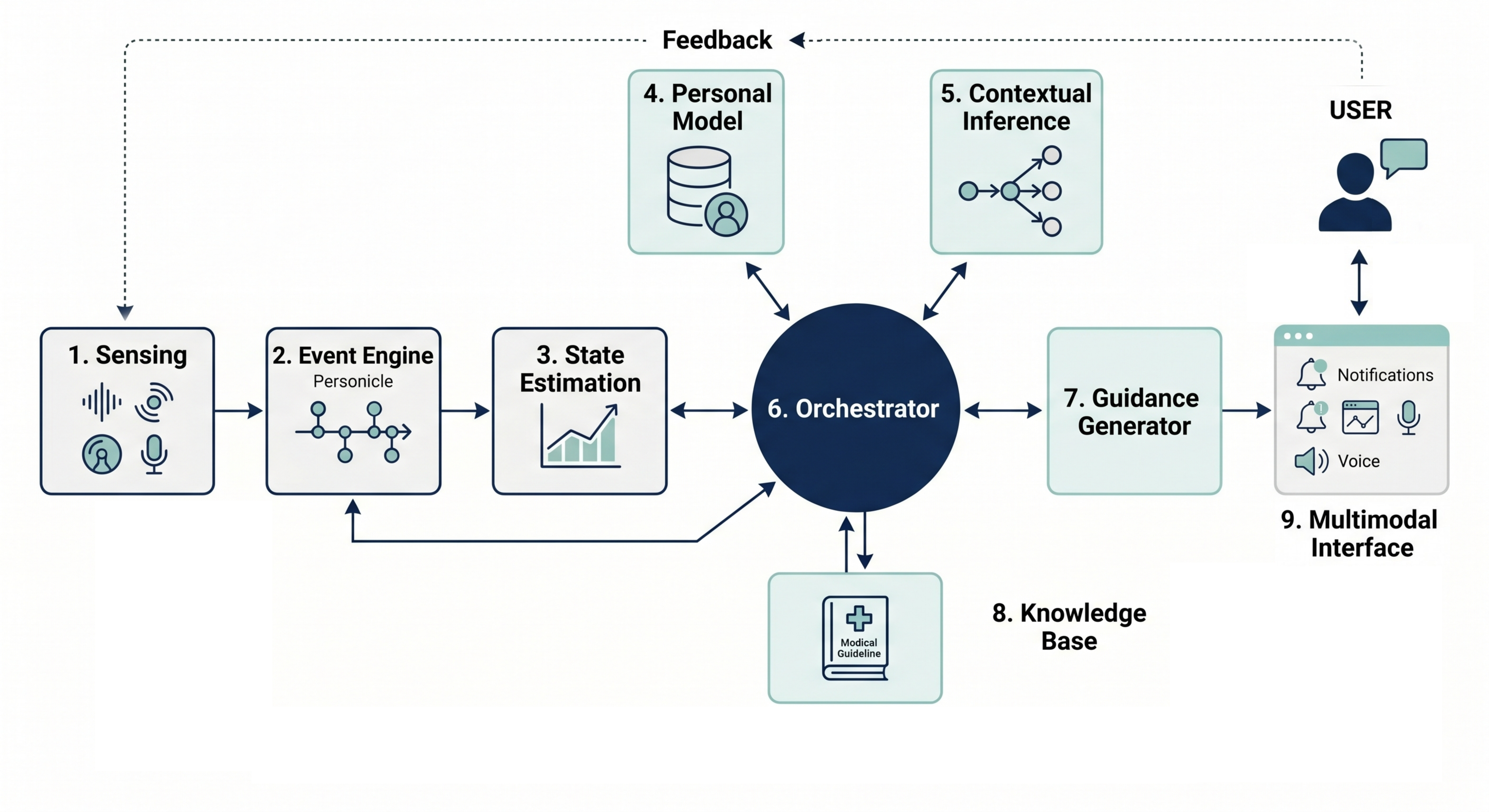} % Spans both columns
    \caption{\textbf{Technical Architecture and Closed-Loop Information Flow of the Personal Health Tech System (PCU).} The diagram illustrates an event-driven system flowing from left to right, centered around a core hub. The sequential ingestion pipeline consists of \textbf{Sensing (1)}, the \textbf{Event Engine / Personicle (2)}, and \textbf{State Estimation (3)}. At the core, the central \textbf{Orchestrator (6)} manages real-time system state via bidirectional communication with three cross-cutting components: the \textbf{Personal Model (4)}, \textbf{Contextual Inference (5)}, and the \textbf{Knowledge Base (8)}. Downstream, the \textbf{Guidance Generator (7)} translates data into actionable insights for the \textbf{Multimodal Interface (9)}. The architecture concludes in a user-centric closed loop, highlighting a bidirectional relationship between the interface and the \textbf{User}, with an overarching, discrete \textbf{Feedback} loop running along the top from the User back to the initial Sensing layer.}
    \label{fig:pcu_architecture}
\end{figure*}

The PCU is a layered, event-driven, agentic architecture for transforming continuous personal data into safe, personalized health guidance. We describe each of its components below. The full architectural specification, including formal definitions and design rationale, is presented in the companion paper~\cite{abbasian2025pcu}. Figure~\ref{fig:pcu_architecture} shows the orchestrator-centric structure, with numbered components corresponding to the subsections that follow.

\subsection{Sensing}

The \textsc{Sensing} layer ingests whatever data a person's environment makes available about their health and daily life: objective physiological measurements (e.g., wearables, CGMs, or other biosensors~\cite{ada2025,battelino2019tir}), subjective inputs such as self-reported symptoms or mood, passive signals from smartphones, ambient devices, or voice~\cite{insel2017digital,powell2024biomarkers}, conversational data, and clinical records. The specific modalities are illustrative, not fixed---new sensor categories can be added as they emerge.

\subsection{Event Engine (Personicle)}

Raw sensor streams are cognitively overwhelming~\cite{jalalijain2021eventmining}; clinicians and patients reason about events, not time series. The \textsc{Event Engine} organizes raw data into a chronological stream of discrete, time-stamped, semantically labeled life events, building on the \textit{Personicle} (Personal Chronicle) abstraction~\cite{jalali2014personicle,jalalijain2021eventmining}. The Personicle represents daily life as a sequence of episodes---``skipped lunch,'' ``walked after dinner,'' ``slept poorly,'' ``missed dose''---identified through interpretable rules and learned patterns~\cite{bulling2014survey}, producing an evolving, computable biography that grounds all downstream reasoning.

\subsection{State Estimation}

The \textsc{State Estimation} module converts events into higher-level estimations of the individual's current health and well-being. Where events are discrete occurrences, a \textit{state} is a latent, continuous construct inferred from multiple events over time: physiological condition, behavioral regularity, emotional trajectory. Crucially, state is measured against the individual's own baseline rather than a population norm---an empirical necessity given the heterogeneity of treatment effects and physiological responses discussed in Section~2. A fasting glucose of 115\,mg/dL within population ``normal'' may be anomalous for someone whose personal baseline is 90; an elevated heart rate unremarkable for one person may signal a meaningful departure for another.

\subsection{Personal Model}

A defining feature of the PCU is that it does not treat each interaction as stateless. The \textsc{Personal Model} is a persistent, structured store of what the system has learned about this specific person: (i)~the individual's longitudinal Personicle and derived behavioral patterns~\cite{jalali2014personicle,jain2014objectiveself}; (ii)~personalized baselines and response curves; (iii)~inferred individual causal patterns---which interventions have reliably attenuated which kinds of events in this person's history; and (iv)~preferences and constraints regarding modality, timing, tone, and data sharing. The component synthesizes ideas from patient-centric knowledge graphs, multi-scale digital twins, and causal memory architectures (Section~2), together with memory designs for LLM agents~\cite{packer2023memgpt,park2023generative}. The Orchestrator queries and writes to the Personal Model on demand. It is what makes the PCU \textit{personal} rather than merely \textit{adaptive}.

\subsection{Contextual Inference}

The \textsc{Contextual Inference} engine asks \textit{why} a detected event occurred, correlating it with behavioral, temporal, and environmental context drawn from both the recent Personicle and the long-term Personal Model. The output is an explanatory attribution that supports targeted, intervention-relevant guidance---not just ``your glucose is high'' but ``this rise is likely because you didn't move after lunch, and walking has reliably blunted similar spikes for you in the past.''

\subsection{Orchestrator}

The \textsc{Orchestrator} is the central hub of the PCU. It maintains bidirectional connections with every other component---reading events from the Event Engine, querying state from the State Estimation Module, retrieving causal attributions from Contextual Inference, looking up personal history and preferences in the Personal Model, drawing on validated content from the Knowledge Base, and writing back to each as new evidence accrues. As events arrive, it decides when to engage which component and how their outputs should combine into action.

At a high level, the PCU as a whole functions as an AI agent: an entity that perceives (Sensing, Event Engine), remembers (Personal Model), reasons about cause and context (State Estimation, Contextual Inference, Knowledge Base), decides (Orchestrator), and acts (Guidance Generator). The Orchestrator is the core that brings this agent together, using reasoning and agentic methods~\cite{yao2023react,schick2023toolformer,hong2024metagpt} to coordinate the other components into coherent behavior. The full design is detailed in the companion paper~\cite{abbasian2025pcu}.

\subsection{Guidance Generator}

The \textsc{Guidance Generator} composes the user-facing output once the Orchestrator has authorized an action. It translates the authorized clinical content into a natural-language message calibrated to the individual's preferences---tone, modality, language, and register---drawing on the Personal Model for delivery context, and delivers the result to users and caregivers through the system's interfacing layer. User responses then flow back into the Sensing layer as conversational data, closing a feedback loop that lets the PCU learn what kinds of communication produce genuine behavior change for each person over time.

\subsection{Knowledge Base}

The \textsc{Knowledge Base} is the PCU's source of validated, externally-sanctioned content: clinical guidelines, condition-specific care standards, behavior change evidence, and safety protocols. It is organized into tiers reflecting how the content is meant to be used---guideline-level recommendations that inform routine guidance, deterministic protocol language for safety-critical situations, and supporting evidence for explanations---so that downstream components can retrieve the right form of knowledge for the decision at hand. Maintaining knowledge as an explicit, queryable resource rather than relying on what a language model happens to recall is what allows the PCU to keep clinical guidance traceable to a verifiable source and updatable as evidence evolves.

\subsection{Multimodal Interface}

The \textsc{Multimodal Interface} is the layer through which the Guidance Generator's output reaches the person, and through which the person reaches back. Different moments call for different channels: a real-time nudge may be best delivered as a brief push notification or voice prompt, a weekly summary as a visual dashboard, an emotionally sensitive conversation as text or speech with appropriately calibrated tone~\cite{abbasian2024empathy}. The interface is also bidirectional---user replies, voice responses, facial cues, and engagement signals flow back through the Sensing layer as conversational data, allowing the system to learn what kinds of communication produce genuine engagement and behavior change for each individual.

% ================================================================
\section{Putting It to Work: From Components to a Type 2 Diabetes Example}
\begin{figure*}[t]
    \centering
    \includegraphics[width=\textwidth]{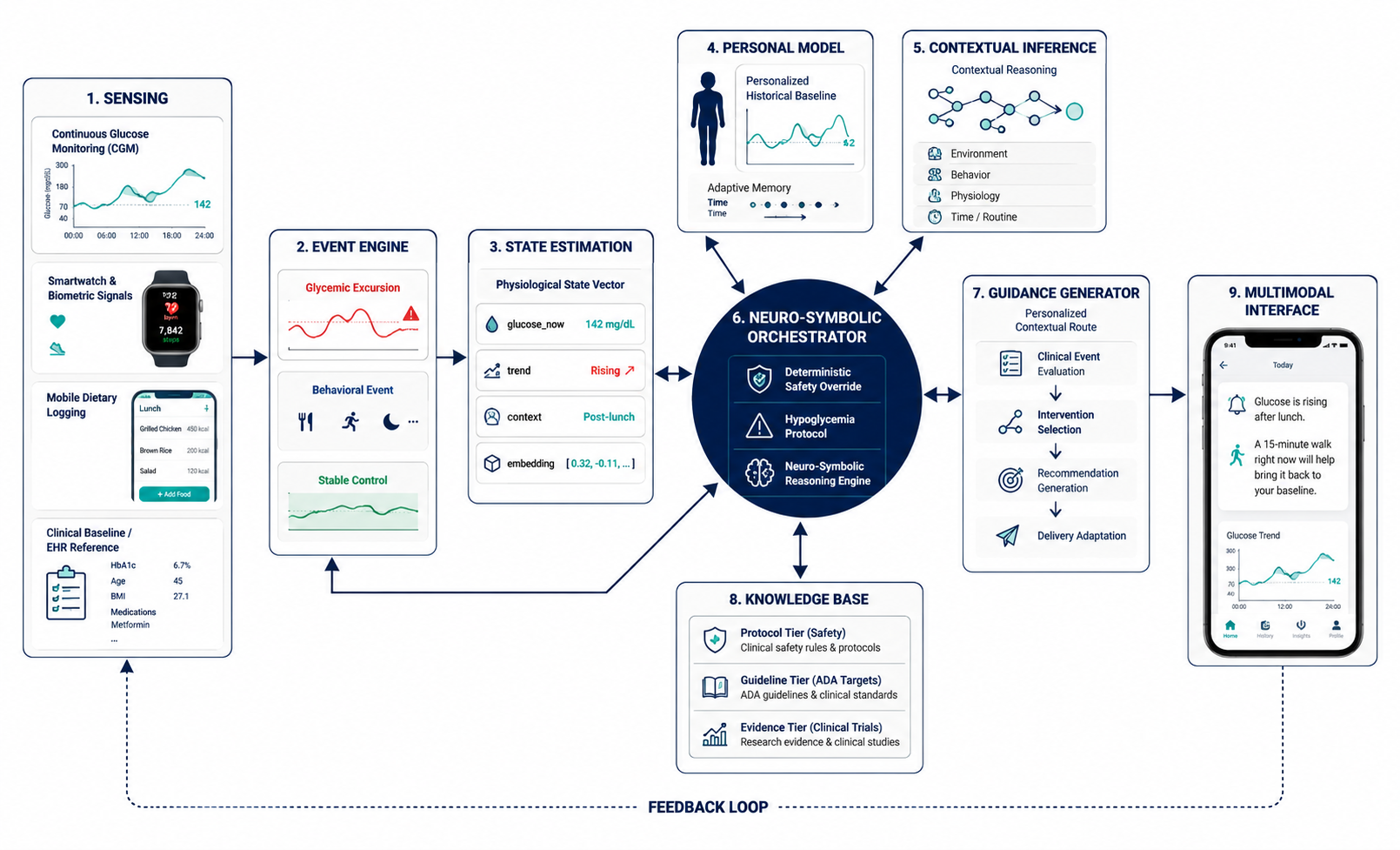}
    \caption{T2D instantiation of the PCU, with representative data at each stage. Sensing (1) ingests CGM, wearable biosignals, dietary logs, and EHR baselines; the Event Engine (2) extracts glycemic excursions, behavioral events, and stable-control states; State Estimation (3) maintains a compact physiological state vector; the Personal Model (4) holds the individual's adaptive baselines and longitudinal history, which Contextual Inference (5) draws on to attribute the current event to its most likely modifiable cause. The neuro-symbolic Orchestrator (6), grounded in a tiered Knowledge Base (8) spanning safety protocols, ADA guidelines, and clinical-trial evidence, routes the event to a deterministic override, a personalized nudge, or silence; the Guidance Generator (7) composes the user-facing message delivered through the multimodal interface (9). The feedback loop updates the Personal Model from observed engagement and outcomes.}
    \label{fig:pcu_t2d}
\end{figure*}

% ================================================================
For every PCU component---Sensing, Event Engine, State Estimation, Personal Model, Contextual Inference, Knowledge Base, Orchestrator, and Guidance Generator---we briefly review the families of methods available in the literature and illustrate one concrete instantiation in the context of Type~2 Diabetes (T2D). T2D is a fitting running example: it affects 589 million adults globally~\cite{idf2025}, is managed almost entirely outside the clinic, is exquisitely sensitive to daily behavioral choices, and is now monitored continuously---the 2025 ADA Standards of Care recommend continuous glucose monitoring for all adults with T2D~\cite{ada2025}, generating 288 glucose readings per day that demand intelligent, \textit{personalized} interpretation. The instantiation below is illustrative, not comprehensive. Figure~\ref{fig:pcu_t2d} shows the complete T2D instantiation we develop across this section

\subsection{Sensing: Multimodal Data Collection}

Sensing in pervasive deployments combines two methodological families. \emph{Passive sensing} streams continuous, objective telemetry from wearables (smartwatches, CGMs) and ambient IoT sensors~\cite{piwek2016rise}; \emph{active sensing} captures subjective or contextual variables---dietary logs, medication confirmations, self-reported stress---through brief user interactions, often via ecological momentary assessment~\cite{intille2012pervasive,klasnja2012healthcare}. Neither is sufficient alone: passive streams miss intent and behaviors invisible to current sensors, while active inputs alone are too sparse to drive real-time guidance. Fusing the two produces a temporally aligned multimodal stream that downstream components can reason over, with enough redundancy that the system degrades gracefully when any one stream is missing or noisy.

\textbf{T2D instantiation.} The AI-READI flagship dataset for T2D illustrates the upper end of what a PCU could ingest---over 15 modalities spanning genomics, EHR, retinal imaging, ECG biosignals, and environmental variables such as home air quality~\cite{AIREADI2024}. For the present illustration, we focus on the signals that most directly drive day-to-day metabolic guidance:

\begin{itemize}
\item \textbf{Continuous Glucose Monitoring (CGM):} the primary physiological anchor, providing interstitial glucose readings every 5 minutes~\cite{sergazinov2024glucobench,marling2020ohiot1dm}.

\item \textbf{Dietary intake:} meal timestamps and estimated composition via user logging or vision models---carbohydrate load is the dominant driver of glycemic excursions~\cite{zeevi2015personalized}.

\item \textbf{Physical activity:} step counts, heart rate, and movement intensity from consumer wearables; combining these with CGM substantially reduces glucose forecasting error~\cite{vandoorn2021maastricht}.

\item \textbf{Medication adherence:} timing and confirmation of metformin, insulin, or other pharmacological interventions, allowing diet-induced and medication-related anomalies to be distinguished.

\item \textbf{Sleep and stress proxies:} sleep duration and autonomic arousal indicators, since both can elevate glucose independently of food intake~\cite{dubosson2018d1namo}.

\item \textbf{Clinical baselines:} longitudinal HbA1c, lipid panels, and metabolic panels from EHR, which anchor the slow-moving disease context~\cite{AIREADI2025}.
\end{itemize}

Table~\ref{tab:t2d_datasets} surveys publicly available T2D datasets combining these signals, varying in cohort size, modalities, and study population.

\begin{table*}[t]
\centering
\footnotesize
\caption{Representative publicly available datasets containing Type 2 diabetes (T2D) CGM or glucose-related data.}
\resizebox{\textwidth}{!}{
% \resizebox{\columnwidth}{!}{
\begin{tabular}{l c c c c}
\hline
\textbf{Dataset} & \textbf{Sample Size} & \textbf{Diabetes Type} & \textbf{Modalities} & \textbf{Population} \\
\hline

AI-READI (2025) \cite{AIREADI2025} 
& Large-scale 
& T2D 
& CGM, EHR, imaging, labs 
& Mixed \\

ShanghaiT2DM (2023) \cite{ShanghaiT2DM} 
& 100 
& T2D 
& CGM, meals, medication, labs 
& Adults \\

Glucose-ML (2025) \cite{GlucoseML2025}
& 2500+ 
& Mixed 
& CGM time-series 
& Multi-country \\

DiaTrend (2023) \cite{DiaTrend2023}
& 54 
& Mixed 
& CGM + insulin pump 
& Adults \\

Broll (2021) \cite{Broll2021}
& 5 
& T2D 
& CGM 
& Adults \\

Colas (2019) \cite{Colas2019}
& 208 (17 T2D) 
& Mixed 
& CGM 
& Adults \\

Hall (2018) \cite{Hall2018}
& 57 
& Mixed 
& CGM + diet 
& Adults \\

Shah (2019) \cite{Shah2019}
& 168 
& Mixed 
& CGM 
& Mixed \\

Ramzi (2019) \cite{Ramazi2019}
& 63 
& T2D 
& CGM + activity + labs 
& Adults \\

\hline
\end{tabular}
}
\label{tab:t2d_datasets}
\end{table*}

\subsection{Event Extraction: From Multimodal Signals to Clinical Episodes}

Approaches to event extraction sit along a spectrum balancing interpretability, flexibility, and safety. \emph{Temporal abstraction}~\cite{shahar1997temporal} and \emph{rule-based} or \emph{complex event processing}~\cite{schneider2022cep} systems map raw values into qualitative states and compose them into clinically meaningful episodes; every event traces to an explicit rule, making them auditable but rigid. \emph{Learned recognition} replaces rules with classifiers and deep sequence models~\cite{bulling2014survey,zhang2022deephar}, gaining flexibility but losing audit trails. \emph{Hierarchical} approaches combine both within a multi-scale architecture~\cite{jalalijain2021eventmining}, detecting primitive events at the signal level and composing them upward into richer episodes---a natural fit for health data, where threshold crossings become excursions and excursions become weekly patterns. \emph{Foundation models} learn latent representations directly from large physiological corpora and have produced strong results in both general wearable sensing~\cite{yuan2024selfsupervised} and condition-specific settings such as continuous glucose monitoring~\cite{lutsker2026foundation}; their main limitation for safety-critical deployment is that auditable chains from detection to recommendation cannot yet be constructed from embeddings alone.

\textbf{T2D instantiation.} For this T2D example, we use a \emph{hierarchical rule-based} approach: the deterministic thresholds from ADA and International Consensus Time-in-Range targets~\cite{ada2025,battelino2019tir} define primitive events at the signal level, and temporal grammars compose them into higher-level clinical episodes. We choose this approach because every event traces to an explicit rule, which is a hard requirement for safety-critical guidance. The same Personicle could be produced by a learned or foundation-model approach, but at the cost of audit chains we are not willing to give up in a column meant to be reproducible.

With this method, the Event Engine produces three classes of events. \emph{Glycemic excursion events}---post-meal spike (glucose rise exceeding a threshold within a defined window after a logged meal), persistent hyperglycemia (sustained elevation by duration), fasting hyperglycemia, hypoglycemia, and severe hypoglycemia (single threshold crossings under context-specific conditions)---feed the safety cascade. Severe hypoglycemia triggers a deterministic protocol that bypasses all downstream reasoning. \emph{Behavior-linked management events}---a missed medication window, an exercise-associated glucose drop---are detected by joining a glycemic primitive with a recent behavioral event from the Personicle. These are the substrate for causal attribution downstream: the system can only explain a deviation in behavioral terms when the behavior is itself a discrete event. \emph{Stable control states} are detected just as deliberately, since silence is the Orchestrator's most frequent output and reliable detection of stability is what prevents over-notification~\cite{nahumshani2018jitai,mishra2021receptivity}.

%#############################################

\subsection{State Estimation: Summarizing What Is Happening Now}

Approaches to state estimation span a spectrum. \emph{Compartmental and mechanistic models} represent physiology through differential equations grounded in biological mechanism---for example, Dalla Man's meal-simulation model of the glucose-insulin system~\cite{dallaman2007}---but require individual parameter calibration that is impractical at scale. \emph{Machine-learning forecasting} operates over windowed feature vectors of recent signals, trading mechanistic interpretability for predictive accuracy and demonstrated across diabetes~\cite{sergazinov2024glucobench,vandoorn2021maastricht} and other continuous-monitoring domains. \emph{Foundation models} learn dense latent representations directly from large physiological corpora and can outperform handcrafted summary statistics~\cite{gluformer2025}. \emph{Meta-learning} learns a population prior that can be rapidly adapted to a new user from limited per-user data~\cite{moon2025bitmaml}. \emph{Compact decision-point representations}, advocated by the JITAI literature, deliberately retain only the small set of features needed to choose among intervention options at the moment of decision~\cite{nahumshani2018jitai,klasnja2015microrandomized}. Richer representations capture more nuance but are harder to audit and more vulnerable to silent failure when an input stream degrades.

\textbf{T2D instantiation.} For T2D, a natural choice is a \emph{compact decision-point representation} in the JITAI style: a small set of fields chosen to be auditable, robust to missing inputs, and sufficient for the next one to two hours of decisions. A PCU built this way could maintain state in two complementary forms. A \emph{physiological state vector} might consist of five fields: \texttt{glucose\_now} (most recent CGM reading); \texttt{trend} (\emph{rising}, \emph{falling}, or \emph{stable} over the past 15--30 minutes); \texttt{meal\_recent} and \texttt{activity\_recent} (whether a meal or meaningful movement occurred in the past two hours and 30 minutes respectively); and \texttt{context} (\emph{fasting}, \emph{post-meal}, \emph{overnight}, or \emph{active})~\cite{ada2025,battelino2019tir,erickson2023walk,bellini2022postprandial,buffey2022sitting}. Alongside it, a \emph{persistent clinical state} would encode the slower-moving disease context: progression stage~\cite{weir2004betacell,mechanick2018dbcd}, complication burden (kidney, heart, liver) from EHR~\cite{ada2025}, and biological phenotype~\cite{ahlqvist2018subgroups}.

The state vector is meant to be interpreted against a \emph{personal baseline} stored in the Personal Model---this individual's typical fasting glucose, postprandial spike magnitude, recovery time, and historical response to activity. Baselines could be initially anchored by clinical safety targets (70--180\,mg/dL and Time in Range above 70\%~\cite{ada2025,battelino2019tir}) and progressively personalized from longitudinal data. This matters because postprandial responses to identical meals vary widely across individuals~\cite{zeevi2015personalized,berry2020predict}: a fasting glucose of 115\,mg/dL within population ``normal'' may be anomalous for someone whose baseline is 90. The persistent clinical state would then adjust how strictly thresholds apply---for example, relaxing the target range under a frailty flag, or raising the salience of medication adherence events when renal progression is high. Under sparse sensing, the vector narrows to populated fields and uncertainty widens accordingly.

\subsection{Contextual Inference: Explaining Why}

Approaches to contextual inference sit along a spectrum from rigid auditability to flexible reasoning. \emph{Rule-based clinical decision support} encodes guideline thresholds and deterministic safety protocols~\cite{sutton2020cdss,ada2025}; auditable but unable to discriminate between mechanisms that yield numerically similar events. \emph{Predictive forecasting} estimates what will happen next from CGM time series and behavioral covariates~\cite{vandoorn2021maastricht,sergazinov2024glucobench}; useful for anticipation but silent on which modifiable factor produced the event. \emph{Digital twins and temporal causal models} maintain person-specific representations of physiology and behavior, with causal extensions oriented toward intervention rather than prediction alone~\cite{katsoulakis2024dt4h,vallee2026causaldt}. \emph{Personal causal memory architectures} such as REMI attach causal schemas to longitudinal personal memory, grounding recommendations in inferred individual cause--effect patterns rather than similarity alone~\cite{raman2025remi}. \emph{Just-in-time adaptive interventions} contribute the timing and receptivity machinery for deciding when to act on an explanation~\cite{nahumshani2018jitai,klasnja2015microrandomized}, though they typically do not themselves produce the mechanism-level account. \emph{LLM-based personal health agents}~\cite{cosentino2025phllm,merrill2024phia,heydari2025pha} can translate technical state into natural-language explanation, but without structured event memory, validated grounding, and explicit causal constraints, fluent explanations may remain unsupported~\cite{omar2025llm,mehandru2024agentsclinic}.

\textbf{T2D instantiation.} For T2D, a natural choice is to combine three of the families above rather than rely on any one. Rule-based logic could handle the safety boundary---severe hypoglycemia and similar threshold-defined emergencies would bypass contextual reasoning entirely and trigger deterministic protocols grounded in ADA and Time-in-Range standards~\cite{ada2025,battelino2019tir}. For non-emergency events, a personal causal memory in the Personal Model would track which interventions have attenuated which kinds of events for this individual---the difference between ``walking has reliably blunted this person's lunch spikes'' and ``walking has not had a consistent effect for this person.'' Predictive forecasting would supply the recovery trajectory used to estimate how long the current event will persist absent intervention. The three families combine into a single attribution object passed to the Orchestrator.

The attribution itself can be organized along three causal classes. \emph{Immediate causes} explain local event dynamics: meal composition, absent post-meal movement, delayed medication dose. \emph{Mediated causes} are contextual modifiers that alter expected intervention effectiveness: poor sleep, acute stress, illness, or circadian disruption. \emph{Accumulated causes} represent longer-term drift: chronic inactivity, repeated nonadherence, or progressively worsening recovery patterns. A useful design property is that causal weights are treated as evolving hypotheses rather than fixed rules---post-meal walking may reliably attenuate excursions during ordinary weeks but lose effectiveness during periods of poor sleep or stress, and the system can update the relevant weights from longitudinal Personicle evidence rather than assuming intervention effects are constant.

%#######################%
\subsection{Orchestrator and Guidance Generator}
\label{sec:kb}

Approaches to orchestration span a spectrum from rigid determinism to unbounded generation, each making a different tradeoff between safety and adaptability. \emph{Rule-based pipelines and finite state machines} map inputs to predefined outputs through explicit logic trees, making every decision fully traceable~\cite{sutton2020cdss}; reliable under known conditions but unable to adapt to individual context or resolve competing goals outside their pre-programmed vocabulary. \emph{Static prompt chaining} embeds language models into a fixed sequence of calls~\cite{wu2022promptchainer}, improving the naturalness of guidance delivery but unable to respond to changing physiological state, behavioral history, or contextual constraints. \emph{Autonomous LLM agents} grant the language model planning authority to evaluate context, invoke tools dynamically, and synthesize a response~\cite{yao2023react,abbasian2025opencha}; flexible across many scenarios, but when the model constructs both the clinical reasoning and the guidance within the same inference step the audit chain from recommendation to evidence becomes difficult to reconstruct~\cite{omar2025llm,mehandru2024agentsclinic}. \emph{Neuro-symbolic architectures} resolve this tension by structurally separating deterministic logic from neural generation~\cite{sarker2021neurosymbolic}: symbolic components govern routing, safety enforcement, and clinical decision-making, while neural components handle pattern recognition and language. Auditability and adaptability coexist because the parts that need to be inspected are inspectable, and the parts that need to be fluent are fluent.

\textbf{T2D instantiation.} For T2D, a natural choice is a neuro-symbolic orchestrator coordinated with a separate guidance generator. The Orchestrator could resolve each incoming event into one of four routing outcomes---safety alert, response to a user question, nudge, or silence---by consulting the Knowledge Base and the Personal Model~\cite{nahumshani2018jitai,wang2025moe,strong2025guideddeferral}. Severe hypoglycemia and similar emergencies would bypass everything downstream and trigger deterministic protocol language; non-urgent actionable events would authorize a nudge with a communication strategy drawn from behavior-change evidence and the individual's longitudinal engagement history~\cite{michie2013bct}; and low-priority contexts would result in silence. The Guidance Generator would then take the authorized clinical content and the selected communication strategy as fixed inputs and compose the user-facing message, calibrated to the individual's preferences stored in the Personal Model.

\subsection{Guidance Generator}

Approaches to guidance generation span a spectrum from classical recommendation to generative reasoning. \emph{Content-based filtering} draws on the individual's own preferences and history to issue suggestions~\cite{basilico2004unifying}; personalized by construction but limited by what the individual has previously engaged with. \emph{Collaborative filtering} recommends what similar users have responded to~\cite{ko2022survey}; powerful at scale but blind to individual physiology. \emph{Hybrid recommenders} combine the two, typically using collaborative methods to handle cold-start and content-based methods to refine over time~\cite{geetha2018hybrid}. \emph{Digital nudging} layers behavior-change framing on top of any recommender, shaping how a suggestion is presented to make action more likely~\cite{jesse2021digital,michie2013bct}. \emph{Retrieval-augmented generation and LLM-based recommenders} use language models to ground suggestions in retrieved evidence and explain them in natural language~\cite{di2023retrieval,peng2025survey}; flexible and conversational, but without explicit grounding they remain vulnerable to fluent-but-unsupported guidance. The common limitation across these families is that they generate suggestions but do not separate clinical content from communication strategy from natural-language expression---when the same component does all three, audit chains and personalization tradeoffs collide.

\textbf{T2D instantiation.} For T2D, a natural choice is a generator that takes the authorized clinical content and a chosen communication strategy as fixed inputs from the Orchestrator and composes only the user-facing message. Personalization happens along three dimensions: \emph{modality} (push notification, weekly digest, voice prompt), \emph{tone and register} (encouraging, factual, plainspoken---learned from the individual's engagement history), and \emph{cultural and linguistic adaptation}~\cite{saenz2025personalized}. Safety alerts bypass all of this---in severe hypoglycemia or similar emergencies the message is deterministic protocol language regardless of preference, modality, or register. This is the architectural invariant the Section as a whole turns on: personalization shapes everything except safety.

\subsection{Knowledge Base}

Approaches to grounding clinical guidance in validated sources span a spectrum. \emph{Curated terminologies and structured databases} such as UMLS~\cite{bodenreider2004umls}, SNOMED CT, and LOINC encode clinical vocabulary in a queryable form, while specialty databases such as DrugBank~\cite{wishart2018drugbank} and SIDER~\cite{kuhn2016sider} cover pharmacological knowledge. \emph{Behavior change taxonomies}~\cite{michie2013bct} provide the corresponding vocabulary for intervention design. \emph{Retrieval-augmented generation} grounds language-model output in retrieved evidence rather than memorized training data~\cite{lewis2020rag,asai2024selfrag}, with clinical variants such as Almanac demonstrating the pattern in medicine~\cite{zakka2024almanac}. The shared limitation is that each source operates in isolation; no single repository covers the guideline, behavioral, and pharmacological knowledge an everyday-health system requires simultaneously.

\textbf{T2D instantiation.} For T2D, a natural choice is a tiered knowledge base. A \emph{guideline tier} would carry ADA Standards of Care~\cite{ada2025} and International Consensus Time-in-Range targets~\cite{battelino2019tir} for clinical thresholds; a \emph{protocol tier} would hold deterministic safety language for emergencies such as severe hypoglycemia; and an \emph{evidence tier} would carry the post-meal-walking trials~\cite{erickson2023walk,bellini2022postprandial,buffey2022sitting} and similar behavior-specific evidence that downstream components draw on. Every recommendation traces back to a tier and a source.

%===============A DAY IN LIFE

\subsection{Putting the Pipeline Together: A Single Question, Two Systems}
\label{sec:dayinlife}

To make the architectural contribution concrete, we follow a single everyday question through both a conventional LLM-based health assistant and the PCU. The question---\emph{``Why is my glucose so high after lunch, and what should I do right now?''}---is representative of the kind of reasoning the 8{,}759 hours outside the clinic demand: the user has a measurement, a context (lunch, the present moment), and a decision to make. Table~\ref{tab:answer_comparison} shows the two responses side by side. Table~\ref{tab:component_attribution} then maps each part of the PCU response to the component that generated it; the same flow is visualized end-to-end in Figure~\ref{fig:pcu_t2d}.

% [TABLE: ANSWER COMPARISON]

The conventional answer is fluent but ungrounded. The model has no access to this person's recent meals, activity, sleep, medication timing, or longitudinal response patterns, and no validated source of clinical evidence beyond what it happened to memorize during training. It produces general advice that could apply to anyone, and the user is left to translate that advice into an actual decision for themselves. The PCU answer, by contrast, is assembled from contributions of multiple components, each grounded in either a deterministic pipeline or a validated source: the State Estimation Module knows that glucose is \emph{rising}, not just elevated; the Event Engine knows that this is the post-meal window of a 62\,g lunch followed by 30 minutes of inactivity; the Personal Model knows that a short walk has historically attenuated similar spikes for this individual; Contextual Inference attributes the rise to the most plausible modifiable factor; the Orchestrator authorizes a low-burden nudge rather than an alert; the Guidance Generator composes the message in the person's preferred tone; and the Confidence \& Safety layer moderates today's recommendation because poor sleep last night may reduce the walk's effectiveness. The answer is not longer than the LLM answer---it is structurally different.

A final architectural point. If the same person had asked the same question while CGM was reading 54\,mg/dL and falling, the entire personalization and explanation pipeline would have been bypassed: the response would have been deterministic protocol guidance grounded in ADA and Time-in-Range standards~\cite{ada2025,battelino2019tir}. \emph{Safety is never personalized; everything else can be.}

\begin{table*}[t]
\centering
\caption{Two responses to the same user question. The conventional LLM produces fluent but ungrounded advice; the PCU assembles a response from multiple components, each grounded in the individual's own data, history, and validated clinical evidence.}
\label{tab:answer_comparison}
\scriptsize
\setlength{\tabcolsep}{4pt}
\renewcommand{\arraystretch}{1.2}
\begin{tabularx}{\textwidth}{>{\raggedright\arraybackslash}p{3.2cm} X}
\toprule
\multicolumn{2}{l}{\textbf{User question:} \emph{``Why is my glucose so high after lunch, and what should I do right now?''}} \\
\midrule
\textbf{Conventional LLM assistant} & 
``Post-meal glucose spikes are common in type 2 diabetes and are often related to carbohydrate intake and a lack of physical activity. To bring your levels down, you can try going for a short walk, drinking water, and avoiding additional carbohydrates. If your glucose remains high or you feel unwell, contact your healthcare provider.'' \\
\addlinespace[1mm]
\textbf{PCU response} & 
``Your glucose is rising after lunch---now at 212\,mg/dL. The most likely reason is that you've been sitting since your 62\,g lunch at 12:05, and a short walk has reliably lowered similar spikes for you in the past (about 30\% on average). If you can, try a 10--15 minute walk in the next half hour; I'll check back in around 30 minutes. One note for today: your sleep was only 5 hours last night, which sometimes reduces how much walking helps, so I'll watch the trend more carefully than usual.'' \\
\bottomrule
\end{tabularx}
\end{table*}

\begin{table}[!htbp]
\centering
\caption{Each component of the PCU contributes a distinct part of the response in Table~\ref{tab:answer_comparison}, grounded in either the individual's own data, longitudinal history, or validated clinical evidence---rather than in a single model's memorized training data.}
\label{tab:component_attribution}
\scriptsize
\setlength{\tabcolsep}{3pt}
\renewcommand{\arraystretch}{1.22}
\begin{tabularx}{\columnwidth}{>{\raggedright\arraybackslash}p{2.4cm} X}
\toprule
\textbf{Component} & \textbf{What it contributes, and why} \\
\midrule
State Estimation 
& 
\textbf{``Your glucose is \emph{rising} after lunch---now at 212\,mg/dL.''} The trajectory (\emph{rising}, not just elevated) and the metabolic context (\emph{post-meal}) come from the physiological state vector. A threshold-based system can report the value; only state estimation can report the dynamic. \\

\addlinespace[0.5mm]

Event Engine 
& 
\textbf{``Since your 62\,g lunch at 12:05.''} The lunch carbohydrate load and the post-meal sedentary window are discrete semantic events extracted from the raw sensing stream, not features computed at query time. Without eventization, no downstream component could refer to them by name. \\

\addlinespace[0.5mm]

Personal Model 
& 
\textbf{``A short walk has reliably lowered similar spikes for you in the past (about 30\% on average).''} The historical effectiveness of walking for this individual lives in the Personal Model, accumulated over previous Personicle episodes. A general-purpose LLM cannot make this claim because it has no record of this person's prior responses. \\

\addlinespace[0.5mm]

Contextual Inference 
& 
\textbf{``The most likely reason is that you've been sitting.''} Among several plausible mechanisms for the rise (meal load, missed medication, stress, sleep), Contextual Inference identifies absent post-meal movement as the most plausible modifiable contributor for this person under these conditions---an attribution, not a forecast. \\

\addlinespace[0.5mm]

Orchestrator 
& 
\textbf{Decision to send a low-burden nudge---not a safety alert, not silence---and to follow up in 30 minutes.} The event is actionable but not urgent, so the Orchestrator routes it to a nudge rather than the safety cascade or the silence default, and schedules a follow-up to verify the recommendation worked. \\

\addlinespace[0.5mm]

Guidance Generator 
& 
\textbf{The message's tone, brevity, and check-in offer.} Length, register, and modality are calibrated to preferences stored in the Personal Model. The same clinical content for a user who prefers weekly digests would appear instead in their next summary, not as a real-time push. \\

\addlinespace[0.5mm]

Knowledge Base 
& 
\textbf{The 70--180\,mg/dL target range and the post-meal walking evidence.} The thresholds and the supporting evidence come from validated sources (ADA Standards of Care, Time-in-Range consensus, post-meal walking trials~\cite{ada2025,battelino2019tir,erickson2023walk,bellini2022postprandial})---not from what the language model happened to memorize during training. \\

\addlinespace[0.5mm]

Confidence \& Safety 
& 
\textbf{``Your sleep was only 5 hours last night, which sometimes reduces how much walking helps.''} The short-sleep event from earlier in the day flags reduced confidence in the activity-based intervention. If the same glucose reading had been 54\,mg/dL and falling, this layer would override everything else and trigger deterministic protocol guidance, regardless of personalization. \\

\bottomrule
\end{tabularx}
\end{table}

% % ================================================================

\section{Toward a Personal Health Infrastructure: Generalizability, Governance, and Open Challenges}

The T2D illustration is one instance of a pattern that transfers broadly, but the claim deserves precision. What is condition-agnostic in the PCU is the pipeline itself: the transformation from raw signals to events, from events to personalized state, from state to contextual inference, and from inference to grounded, preference-adapted guidance. What is condition-specific is the vocabulary that fills it---the sensing modalities, the event definitions, the knowledge base content, and the clinical thresholds. Separating these two layers is what makes the architecture reusable rather than merely flexible. An asthma PCU would replace CGM with peak flow meters and environmental sensors and ground its knowledge base in GINA guidelines; a cardiac rehabilitation PCU would monitor exercise tolerance relative to individual recovery trajectories; a mental-health PCU would replace physiological anchors with behavioral markers such as sleep disruption, social withdrawal, and activity decline~\cite{rahmani2022mental}. In each case the events, knowledge, and thresholds change; the orchestration logic, safety cascade, Personal Model structure, and delivery personalization remain identical.

Deployment at scale depends on interoperability with existing health infrastructure. CGM readings map to HL7 FHIR Observation resources, event streams can be encoded using SNOMED CT and LOINC, and SMART on FHIR enables EHR integration~\cite{smartonfhir2024}---making the PCU a layer that connects to the clinical infrastructure already surrounding the person rather than a parallel system. At population scale, aggregating de-identified event streams under federated learning and purpose-bounded sharing~\cite{abbasian2025pcu} creates a substrate for epidemiological research that cross-sectional studies cannot surface. Continuous high-fidelity behavioral and physiological data across diverse populations enables real-time assessment of intervention effectiveness, early detection of equity gaps, and the discovery of new phenotypes that only become visible when daily life is observable. The individual benefit and the population benefit are, in this architecture, the same data pipeline: every event a PCU helps interpret for one person is, with appropriate consent and privacy protections, evidence that helps the next.

\textbf{Safety, scope, and the personalization boundary.} Any always-on personal health agent requires a safety case. The PCU's rests on scope clarity: the system provides education, lifestyle nudges, safety alerts, and escalation to providers; it never diagnoses, prescribes, or adjusts medication doses. Under FDA clinical decision support guidance, software that informs rather than replaces clinical judgment occupies a lower-risk regulatory tier~\cite{fda2026cds}, and the WHO and NIST AI governance frameworks anchor the system's transparency, accountability, and fairness obligations~\cite{who2021ai,nist2023ai}. The most critical invariant is architectural: user preferences govern timing, tone, modality, and content depth, but never clinical safety thresholds. A user who disables notifications still receives hypoglycemia alerts; a user who requests a higher glucose target still triggers the standard safety protocol. The deterministic safety cascade operates independently of the personalization layer and cannot be overridden by any learned preference or explicit configuration. Safety is a hard constraint, not a default setting.

\textbf{Equity, traceability, and evaluation.} Three further governance challenges shape the research agenda. Equity demands explicit design attention: access to continuous monitors, smartphones, and reliable connectivity is unevenly distributed, and digital health tools that assume a full sensor stack risk amplifying disparities rather than reducing them~\cite{veinot2018goodintent,sabben2024access,rodriguez2025langdisp}. The PCU's graceful degradation---functioning usefully with minimal sensors, adapting to literacy level, supporting multiple languages---is an equity argument embedded in the architecture, but evaluation must assess benefit and risk across demographic groups, not only on average. Traceability is harder. As the Personal Model accumulates longitudinal data and progressively personalizes its baselines, causal weights, and delivery preferences, earlier recommendations may no longer be reproducible from the system's current state---a real concern for clinical accountability, since a clinician reviewing a past nudge must be able to understand why it was issued. Requiring every recommendation to trace to a knowledge base source addresses this partially; fully auditing a continuously-evolving Personal Model will likely require versioned snapshots of the model state at the time of each decision, together with decision logs capturing the inputs, the attribution, and the chosen action. Finally, evaluation methodology must catch up to the architecture. Microrandomized trials~\cite{klasnja2015microrandomized} are well-suited to evaluating per-decision personalization within a deployed PCU, while longer-horizon outcomes (HbA1c, time-in-range, hospitalization, equity-stratified benefit) require prospective cohort designs comparing PCU-based care against threshold-based baselines. Neither methodology is unfamiliar; what is new is needing both at once.

These challenges---together with robust multimodal fusion under heterogeneous data quality and scalable personalization with explicit user control over what the system knows and does---constitute a concrete research agenda for the pervasive computing community. The PCU is not a finished system; it is a blueprint for how the pieces should fit, and the substantive work of building, evaluating, and governing the resulting systems lies ahead.

The missing layer in digital health is not more data, not more AI, and not more apps---it is \textit{personal infrastructure}. This vision builds on two decades of intellectual groundwork, from Hood's P4 Medicine~\cite{hood2011p4} to Estrin's ``Small Data, Where N=Me''~\cite{estrin2014smalldata}, from Health Navigation as a cybernetic GPS for well-being~\cite{nag2019health} to the Personicle as a computable chronicle of lived experience~\cite{jalali2014personicle}. The PCU brings these threads together into an architecture where N truly equals one. The 8{,}759 hours will not navigate themselves; but with the right personal infrastructure---event by event, nudge by nudge, calibrated to the person at the center---they can be transformed into computable contexts for trustworthy health guidance. Like all good infrastructure, the PCU's highest achievement will be that it feels less like technology and more like understanding.

% ================================================================
% REFERENCES
% ================================================================
\bibliographystyle{IEEEtran}
\bibliography{references}
\end{document}